\documentclass{article}

\usepackage{microtype}
\usepackage{graphicx}
\usepackage{subfigure}
\usepackage{booktabs} 
\usepackage{xspace}
\usepackage{amsmath,amsfonts,amssymb,amsthm}
\usepackage[dvipsnames]{xcolor}
\usepackage{diagbox}
\usepackage{pifont}
\usepackage{amsmath}
\usepackage{natbib}
\usepackage{enumitem}

\usepackage{hyperref}

\usepackage[accepted]{icml2020}

\icmltitlerunning{Informative Dropout for Robust Representation Learning:
A Shape-bias Perspective}

\begin{document}
\setlength{\textfloatsep}{10pt plus 1.0pt minus 2.0pt}

\twocolumn[

\icmltitle{Informative Dropout for Robust Representation Learning:\\
A Shape-bias Perspective}



\icmlsetsymbol{equal}{*}

\begin{icmlauthorlist}
\icmlauthor{Baifeng Shi}{equal,pku_cs}
\icmlauthor{Dinghuai Zhang}{equal,pku_sms}
\icmlauthor{Qi Dai}{msra}
\icmlauthor{Zhanxing Zhu}{pku_sms,pku_cds,bibdr}
\icmlauthor{Yadong Mu}{pku_wangxuan}
\icmlauthor{Jingdong Wang}{msra}
\end{icmlauthorlist}

\icmlaffiliation{pku_cs}{School of EECS, Peking University, China}
\icmlaffiliation{pku_sms}{School of Mathematical Sciences, Peking University, China}
\icmlaffiliation{msra}{Microsoft Research Asia}
\icmlaffiliation{pku_cds}{Center for Data Science, Peking University}
\icmlaffiliation{bibdr}{Beijing Institute of Big Data Research}
\icmlaffiliation{pku_wangxuan}{Wangxuan Institute of Computer Technology, Peking University}

\icmlcorrespondingauthor{Baifeng Shi}{bfshi@pku.edu.cn}

\icmlkeywords{Texture-bias, Shape-bias, Robustness,
Domain generalization, Few-shot Learning, Image corruption,
Adversarial perturbation,
Convolutional Neural Network, Computer vision, InfoDrop, ICML}

\vskip 0.3in
]



\printAffiliationsAndNotice{\icmlEqualContribution} 



\makeatletter
\DeclareRobustCommand\onedot{\futurelet\@let@token\@onedot}
\def\@onedot{\ifx\@let@token.\else.\null\fi\xspace}

\def\eg{\emph{e.g}\onedot} \def\Eg{\emph{E.g}\onedot}
\def\ie{\emph{i.e}\onedot} \def\Ie{\emph{I.e}\onedot}
\def\cf{\emph{c.f}\onedot} \def\Cf{\emph{C.f}\onedot}
\def\etc{\emph{etc}\onedot} \def\vs{\emph{vs}\onedot}
\def\wrt{w.r.t\onedot} \def\dof{d.o.f\onedot}
\def\etal{\emph{et al}\onedot}
\def\viz{\emph{viz}\onedot}
\makeatother



\newcommand{\cmark}{\ding{51}}%
\newcommand{\xmark}{\ding{55}}%


\begin{abstract}
Convolutional Neural Networks (CNNs) are known to rely more on local texture rather than global shape when making decisions. Recent work also indicates a close relationship between CNN's texture-bias and its robustness against distribution shift, adversarial perturbation, random corruption, \etc. In this work, we attempt at improving various kinds of robustness universally by alleviating CNN's texture bias. With inspiration from the human visual system, we propose a light-weight model-agnostic method, namely Informative Dropout (InfoDrop), to improve interpretability and reduce texture bias. Specifically, we discriminate texture from shape based on local self-information in an image, and adopt a Dropout-like algorithm to decorrelate the model output from the local texture. Through extensive experiments, we observe enhanced robustness under various scenarios (domain generalization, few-shot classification, image corruption, and adversarial perturbation). To the best of our knowledge, this work is one of the earliest attempts to improve different kinds of robustness in a unified model, shedding new light on the relationship between shape-bias and robustness, also on new approaches to trustworthy machine learning algorithms. Code is available at \url{https://github.com/bfshi/InfoDrop}.
\end{abstract}


\section{Introduction}
\label{introduction}

\begin{figure}[ht]
\begin{center}
\centerline{\includegraphics[width=\columnwidth]{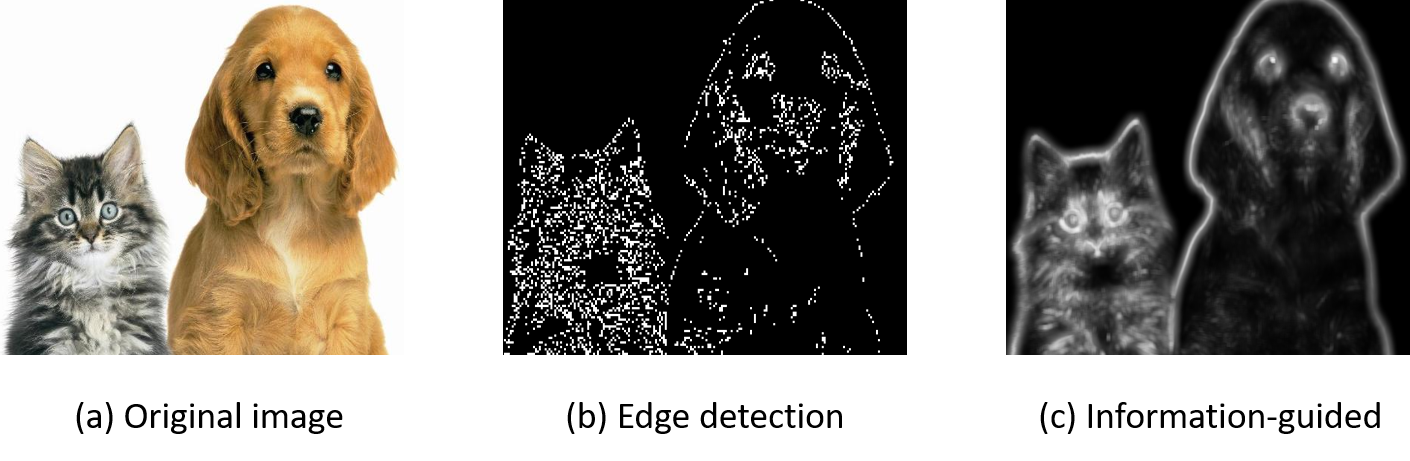}}
\vskip -0.13in
\caption{Comparison of different shape-biased methods. (a) Original image of cat and dog. (b) Simple edge detection is susceptible to complex patterns (\eg stripes of the cat) and can severely damage image contents. (c) In this work, we reduce texture-bias under guidance of self-information, which aligns well with human vision.  The definition and computation of self-information are in Sec. \ref{sec: method}.}
\label{intro_fig}
\end{center}
\vskip -0.23in
\end{figure}

Despite the impressive performance in a broad range of visual tasks, Convolutional Neural Network (CNN) is surprisingly vulnerable compared with the human visual system.
For example, features learned by CNN have trouble in generalizing across shifted distributions between training and test data~\cite{chen2018a,wang2019learning}. Random image corruptions can also considerably degrade its performance~\cite{hendrycks2019benchmarking}. CNN is extremely defenseless under well-designed image perturbation as well~\cite{szegedy2013intriguing}.
This is opposite to the human visual system, which is robust to domain gap, noisy input, \etc~\cite{biederman1987recognition,bisanz2012learning,geirhos2017comparing}.

Another intriguing property of CNN is its `texture bias', namely its bias towards texture instead of shape. Despite the earlier belief that CNN extracts more abstract shapes and structures layer by layer as human does~\cite{kriegeskorte2015deep,lecun2015deep}, recent works reveal its reliance on the local texture when making decisions~\cite{geirhos2018imagenet,brendel2019approximating}. For instance, given an image with a cat's shape filled with an elephant's skin texture, CNN tends to classify it as an elephant instead of a cat~\cite{geirhos2018imagenet}.

Supported by some recent works, there seems to be a surprisingly close relationship between CNN's robustness and texture-bias. For example, \citet{zhang2019interpreting} find that adversarially trained CNNs are innately less texture-biased. There are also a few attempts to tackle a specific task by training a less texture-biased model. \citet{carlucci2019domain} propose to improve robustness against domain gap by training on jigsaw puzzles, which relies more on global structure information. \citet{geirhos2018imagenet} find that shape-biased CNNs trained on stylized images are more robust to random image distortions. Up to this point, one may naturally wonder:

\emph{Is texture-bias a common reason for CNN's different kinds of non-robustness against distribution shift, adversarial perturbation, image corruption, etc.?}

To explore the answer, this work aims at improving various kinds of robustness universally by alleviating CNN's texture bias and enhancing shape-bias. Some approaches to train shape-biased CNNs have been proposed recently. However, they either are susceptible to complex patterns (see Fig.~\ref{intro_fig}(b))~\cite{radenovic2018deep}, or have high computational complexity and auxiliary tasks~\cite{geirhos2018imagenet,wang2019learning,carlucci2019domain,wang2018learning}. In this work, we propose a light-weight model-agnostic method, namely Informative Dropout (\textbf{InfoDrop}). The inspiration comes~from earlier works on saliency detection and human eye movements: humans tend to look at regions with high self-information $-\log\mathbb{P}(\text{current region} \ | \  \text{surrounding regions})$, \ie, regions whose being observed based on surrounding regions contains more `surprise'~\cite{bruce2006saliency,bruce2009saliency}. In other words, people tend to pay more attention to regions that look different from neighboring regions. In our case, patterns like flat regions or high-frequency textures tend to repeat themselves in the neighboring region, thus being less informative. On the other hand, visual primitives (\eg edges, corners) are more unique and thus more informative among its neighborhood. Fig.~\ref{intro_fig}(c) provides a visualization of the information distribution in natural images. Note that both shape and important characteristics (\eg eyes, stripes) are accentuated, while texture (\eg hair) is relatively repressed.

To this end, InfoDrop is proposed to reduce texture-bias by decorrelating each layer's output with less informative input regions. Specifically, we adopt a Dropout-like algorithm~\cite{srivastava2014dropout}: for input regions with less information, we zero out the corresponding output neurons with higher probability. In this way, reliance on textures can be reduced and the model is trained to be more biased towards shapes.
By eliminating InfoDrop after training, the model is further demonstrated to be internally shape-biased without InfoDrop during inference.
The shape-bias property is exhibited through different experiments, both qualitatively and quantitatively.

To evaluate the robustness of InfoDrop, we conduct extensive experiments in four different tasks: domain generalization, few-shot classification, robustness against random corruption, and adversarial robustness. Results show a consistent improvement in different kinds of robustness over various baselines, demonstrating the effectiveness and versatility of our method. We also demonstrate that InfoDrop can be combined with other algorithms (\eg adversarial training) to further enhance the robustness.

\subsection{Our Contribution}

\begin{itemize}
    \setlength{\itemsep}{-1.1pt}
    \item With inspiration from the human visual system, we propose InfoDrop, an effective albeit simple plug-in method to reduce the general texture bias of any CNN-based model.
    \item As shown by extensive experiments, InfoDrop achieves consistently non-trivial improvement over multiple baselines in a wide variety of robustness settings. Furthermore, InfoDrop can be incorporated together with other algorithms to obtain higher robustness.
    \item To the best of our knowledge, this work is one of the earliest attempts to improve different kinds of robustness in a unified model. This sheds new light on the relationship between CNN's texture-bias and non-robustness, also on new approaches to building trustworthy machine learning algorithms.
\end{itemize}


\section{Related Work}

\subsection{{Vulnerability} of CNNs}

An important feature of intelligence is its ability to generalize knowledge across tasks, domains and categories~\cite{csurka2017domain}. However, CNNs still struggle when different kinds of distribution shifts exist between training and test data. For instance, in few-shot classification, where large class gap is the main challenge, complex algorithms make little improvement upon simple baselines~\cite{chen2018a,huang2020are,dhillon2020a}. CNNs also have trouble with transferring knowledge across different domains, especially when data is unavailable in the target domain as in the task of domain generalization~\cite{khosla2012undoing,li2017deeper,li2018domain,carlucci2019domain}. In this work, we evaluate our method's robustness against distribution shift on tasks of few-shot classification and domain generalization.

CNNs are also sensitive to small perturbations and corruptions in images, which can be easily dealt with by humans~\cite{azulay2018deep}. \citet{hendrycks2019benchmarking} benchmark CNN's robustness against 18 types of random corruption, demonstrating its vulnerability. It is also shown that well-designed perturbation, namely adversarial perturbation, can severely degrade the performance of CNNs~\cite{szegedy2013intriguing}. We evaluate the robustness of our approach against both random corruption and adversarial perturbation, with other methods towards model robustness as baseline, \eg, adversarial training~\cite{madry2017towards, zhang2019you}.

\subsection{Texture-bias of CNNs}

Despite the recent impressive performance of CNNs in various vision tasks, the visual processing mechanism behind remains controversial. One widely accepted hypothesis is that CNNs extract low-level primitives (\eg edges, corners) in lower layers and try to combine them into complex shapes in higher layers~\cite{kriegeskorte2015deep,lecun2015deep}. This hypothesis is supported by numbers of empirical findings, both from computational~\cite{zeiler2014visualizing} and psychological~\cite{ritter2017cognitive} perspectives. However, recent work argues that local texture is sufficient for CNNs to perform correct classification~\cite{brendel2019approximating}. Shape or contour information, on the other hand, seems hard for CNNs to understand~\cite{ballester2016performance}. CNNs also fail at transferring between images with similar shapes yet distinct textures~\cite{geirhos2018imagenet}. These findings indicate an alternative explanation for the success of CNNs: local texture is what CNNs base on when making decisions.

\subsection{Relation between Non-robustness and Texture-bias}

More and more work indicates a close relationship between CNNs' non-robustness and texture-bias. \citet{zhang2019interpreting} find that adversarially trained networks are less texture-biased. \citet{geirhos2018imagenet} show that shape-biased models trained with stylized images are more robust against image distortion. \citet{carlucci2019domain} propose to boost domain generalization by training to solve jigsaw puzzles, which relies more on global structure. \citet{wang2019learning} propose to penalize CNN's local predictive power to reduce the domain gap induced by image background. With the same objective, \citet{wang2018learning} propose to project out superficial statistics in feature space. However, none of the work has discussed the relationship between texture-bias and different types of non-robustness in a unified model.

\subsection{Bias in Human Vision}

It is known that human eyes tend to fixate on specific regions (saliency) rather than scan the whole image they see~\cite{yarbus2013eye}. The mechanism behind this kind of bias has attracted lots of interest. \citet{itti1998model} reveal the importance of center-surround contrast of units in the human visual system. \citet{hou2007saliency} detect saliency using residual contrast in the spectral domain. Other works propose to use Shannon entropy to measure saliency and predict fixation~\cite{fritz2004attentive,renninger2005information}. In~\citet{bruce2006saliency}, self-information is proposed to  better model saliency.

In addition, shape-bias is also found critical in the human visual system. A large amount of evidence shows shape is the most important single clue for human vision learning and processing~\cite{landau1988importance}. For example, young children tend to extend object names based on its shape, rather than size, color or material~\cite{diesendruck2003specific}. The shape bias of human vision, together with its bias towards self-information, further motivates our proposed method.


\section{Methodology}
\label{sec: method}

Let $\mathbf{x} \in \mathbb{R}^{c_0 \times h_0 \times w_0}$ denotes an image with $c_0$ channels and spatial shape of $h_0 \times w_0$. For a CNN, we denote the input of $l$-th convolutional layer by $\mathbf{z}^{\ell-1} \in \mathbb{R}^{c_{\ell-1} \times h_{\ell-1} \times w_{\ell-1}}$ and output by $\mathbf{z}^\ell \in \mathbb{R}^{c_\ell \times h_\ell \times w_\ell}$. Note that $\mathbf{z}^0$ equals to the input image $\mathbf{x}$. Assume the $l$-th layer has a convolutional kernel $\mathbf{k}^\ell \in \mathbb{R}^{c_\ell \times c_{\ell-1} \times k \times k}$ and bias $\mathbf{b}^\ell \in \mathbb{R}^{c_\ell}$, where $k$ is the kernel size. Then for $c$-th channel's $j$-th element $z_{c,j}^\ell$ in output $\mathbf{z}^\ell$ ($j\in \{1, 2, ..., h_\ell w_\ell\}$), we have $z_{c,j}^\ell = \sigma(\mathbf{k}_c^\ell \cdot \mathbf{p}_j^{\ell-1} + b_c^\ell)$, where $\mathbf{p}_j^{\ell-1} \in \mathbb{R}^{c_{\ell-1} \times k \times k}$ is the $j$-th patch in $\mathbf{z}^{\ell-1}$, $\mathbf{k}_c^\ell$ and $b_c^\ell$ are the kernel and bias for $c$-th output channel, $\cdot$ indicates inner product and $\sigma(\cdot)$ is an entry-wise activation function (\eg ReLU). All through this paper $\Vert\cdot\Vert$ denotes Euclidean norm.

\subsection{Informative Dropout}
Now we develop our information-based Dropout method for alleviating texture-bias. As discussed in Section~\ref{introduction}, regions of textures tend to contain low self-information. To this end, we propose to reduce texture-bias by decorrelating each layer's output with low-information regions in input. Specifically, we adopt a Dropout-like approach for the purpose. In traditional Dropout \cite{srivastava2014dropout}, a multiplicative Bernoulli noise is introduced to help prevent overfitting, where each neuron is zeroed out with equal probability. In order to suppress texture-bias, we propose to zero out an output neuron with higher probability if the input patch contains less information, and vice versa. Specifically, we model the drop coefficient $\textit{r}$ of the $j$-th neuron in output's $c$-th channel with a Boltzmann distribution:

\vspace{-0.5em}
\begin{equation}
\label{sample_prob}
    \textit{r}(z^\ell_{c,j}) \propto e^{- {\mathcal{I}}(\mathbf{p}^{\ell-1}_j) / T},
\end{equation}
where $\mathbf{p}^{\ell-1}_j$ is the patch in the input related to the computation of $z^\ell_{c,j}$, $\mathcal{I}$ denotes self-information and $T$ is temperature. When value of $\mathcal{I}$ is low, the corresponding neuron is likely to be dropped, and the network tends to rely less on $\mathbf{p}^{\ell-1}_j$. Here the temperature $T$ serves as a `soft threshold' of information. When $T$ is small, the threshold lowers down, and only patches with least information (\eg a patch in a solid-colored region) will be dropped. When $T$ goes to infinity, all neuron will be dropped with equal probability, and the whole algorithm becomes regular Dropout.

First we discuss how to estimate $\mathcal{I}$. The definition of information could date back to Shannon's work~\cite{shannon1948mathematical}, from where we borrow the concept of self-information $\mathcal{I}$ to describe the information of a patch:

\vspace{-0.5em}
\begin{equation}
    \mathcal{I}(\mathbf{p}^{\ell-1}_j) = -\log \textit{q}^{\ell-1}_j (\mathbf{p}^{\ell-1}_j),
\end{equation}

where $\textit{q}^{\ell-1}_j$ is the distribution which $\mathbf{p}^{\ell-1}_j$ is sampled from, if we see $\mathbf{p}^{\ell-1}_j$ as a realization of a random variable. As a simple case, we can assume that all patches in the neighborhood of $\mathbf{p}^{\ell-1}_j$ are different realizations of the same random variable, \ie, they are all sampled from the same distribution $\textit{q}^{\ell-1}_j$. In this case, if $\mathbf{p}^{\ell-1}_j$ contains more ``texture" than ``shape", its pattern shall repeat itself within a local region, resulting in a high likelihood $\textit{q}^{\ell-1}_j (\mathbf{p}^{\ell-1}_j)$ and hence low self-information and should be zeroed out with high probability.

To approximate $\textit{q}^{\ell-1}_j(\cdot)$, we assume that $\mathbf{p}^{\ell-1}_j$ and other patches in its neighbourhood $\mathcal{N}^{\ell-1}_j$ come from the same distribution $\mathbf{p} \sim \textit{q}^{\ell-1}_j(\mathbf{p})$. Here the neighbourhood means a local region centered at $\mathbf{p}^{\ell-1}_j$, with Manhattan radius $R$, \ie, the neighborhood contains $(2R + 1)^2$ patches. Then, with neighboring patches as samples\footnote{Here all the patches in the neighborhood are used. Nonetheless, one can only use a random part of the patches for an unbiased estimation to reduce the computational load, especially when the radius of the neighborhood is large. From our observation, this barely affects the performance.}, we approximate $\textit{q}^{\ell-1}_j(\cdot)$ with its kernel density estimator $\hat{\textit{q}}_j^{\ell-1}$, \ie



\begin{equation}
\label{patch_distribution}
     \hat{\textit{q}}_j^{\ell-1}(\mathbf{p}) = \frac{1}{(2R + 1)^2} \sum_{\mathbf{p}^\prime \in \mathcal{N}^{\ell-1}_j} \textit{K}(\mathbf{p}, \mathbf{p}^\prime),
\end{equation}
where $\textit{K}(\cdot, \cdot)$ is kernel function. Here we use Gaussian kernel, \ie, $\textit{K}(\mathbf{p}, \mathbf{p}^\prime) = \frac{1}{\sqrt{2\pi}h} \exp(-||\mathbf{p} - \mathbf{p}^\prime||^2 / 2h^2)$, where $h$ is the bandwidth. Then the information of $\mathbf{p}^{\ell-1}_j$ is estimated by
\begin{equation}\small
     \mathcal{\hat{I}}(\mathbf{p}^{\ell-1}_j) = -\log \{\sum_{\mathbf{p}^\prime \in \mathcal{N}^{\ell-1}_j} e^{-||\mathbf{p}^{\ell-1}_j - \mathbf{p}^\prime||^2 / 2h^2}\} \ + \ \textmd{const}.
\end{equation}
As one can observe, the more different $\mathbf{p}^{\ell-1}_j$ is from neighbouring patches, the more information it contains. For regions of solid color or high-frequency texture, similar patterns tend to repeat in the neighborhood, and thus little information is presented. Local shapes, on the other hand, are more unique in their surroundings and thus more informative.

Then we discuss how the dropout process works. A direct way is to sample neurons in the output $\mathbf{z}^\ell$ with probabilities given by Eq.~\ref{sample_prob}, and set them to zero. During training, for the $c$-th channel of $\ell$-th layer's output $\mathbf{z}_c^\ell \in \mathbb{R}^{h_\ell \times w_\ell}$, we randomly choose neurons to drop by running weighted multinomial sampling with replacement for $r_0 \cdot h_{\ell} \cdot w_{\ell}$ times,\footnote{Here we choose sampling with replacement over without replacement because the former runs faster in practice. Hence here $r_0$ can be any positive real number due to collision of samples, and the actual dropout rate (expected ratio of sampled neurons) will be lower than $r_0$.} where $r_0$ is a hyper-parameter controlling the amount of dropped neurons. The algorithm is shown in  Alg.~\ref{alg1}.

Note that when training with InfoDrop on, we are \emph{intentionally} filtering out texture to make the model learn to recognize shape. However, during inference, we expect to see a genuinely shape-biased model which can filter out texture by itself \emph{without} InfoDrop's help. To check if our model has obtained this ``internal'' shape-bias, one way is to directly remove the InfoDrop blocks during inference. However, there may be statistical mismatch (\eg in batch normalization) between clean images and images processed by InfoDrop. To this end, we take the inspiration from~\cite{geirhos2018imagenet} and propose to \textbf{finetune the network on clean images with InfoDrop removed}, as an extra step after InfoDrop training. In this way, we can safely remove InfoDrop during testing, and examine whether our network has truly learned shape-bias.


\begin{algorithm}[tb]
   \caption{Informative Dropout (InfoDrop)}
   \label{alg1}
\begin{algorithmic}
   \STATE {\bfseries Input:} input activation map $\mathbf{z}^{\ell-1}$
   \STATE {\bfseries Parameters:} convolutional kernel $\mathbf{k}^\ell$, bias $\mathbf{b}^\ell$, radius $R$, temperature $T$, bandwidth $h$, ``dropout rate'' $r_0$
   \STATE {\bfseries Output:} output activation map $\mathbf{z}^\ell$
   \\~\\
   \FOR{each element $z^\ell_{c,j}$ in output}
   \STATE $z^\ell_{c,j} \gets \sigma(\mathbf{k}_c^\ell \cdot \mathbf{p}^{\ell-1}_j + b_c^\ell)$
   \ENDFOR

   \FOR{$c = 1$ {\bfseries to} $c_\ell$}
   \FOR{$i = 1$ {\bfseries to} $\lfloor r_0 \cdot h_\ell \cdot w_\ell \rfloor$}
   \STATE sample $j$ from $[1, h_\ell \cdot w_\ell]$ with probability $\textit{r}(z^\ell_{c,j})$ given by Eq.~\ref{sample_prob}
   \STATE $z^\ell_{c,j} \gets 0$
   \ENDFOR
   \ENDFOR

\end{algorithmic}
\end{algorithm}

\subsection{Computational Complexity}

There are two parts of computational cost in InfoDrop: (i) calculation of self-information for input patches, and (ii) manipulation of each output element. For self-information calculation, there are $\mathcal{O}(h_{\ell-1} \cdot w_{\ell-1})$ input patches, each with size $\mathcal{O}(c_{\ell-1})$. Note that kernel size and scale of neighborhood are constants. This means a time complexity of $\mathcal{O}(c_{\ell-1} \cdot h_{\ell-1} \cdot w_{\ell-1})$ for part (i). As for part (ii), both sampling and element-wise product needs $\mathcal{O}(c_\ell \cdot h_\ell \cdot w_\ell)$. Note that spatial shape often stays unchanged through convolution. Therefore, time complexity of InfoDrop is $\mathcal{O}((c_{\ell-1} + c_\ell) \cdot h_\ell \cdot w_\ell)$, which is little overhead compared with $\mathcal{O}(c_{\ell-1} \cdot c_\ell \cdot h_\ell \cdot w_\ell)$ in convolutional operation.


\section{Experiments}

We conduct extensive experiments for further understanding properties of InfoDrop and its benefits over standard CNN-based models. First we discuss the shape-bias property of InfoDrop in Sec.~\ref{infodrop_shape_bias}. Then in Sec.~\ref{infodrop_robustness} we evaluate robustness of InfoDrop through four different tasks, \viz domain generalization, few-shot classification, robustness against random corruption and adversarial robustness, and also compare with other shape-biased approaches. In Sec.~\ref{ablation}, we conduct ablation studies for further analysis. The balance between shape and texture is discussed in Sec.~\ref{sec:is_shape_all_we_need}. Please refer to Appendix for specific experimental settings.

\subsection{Shape-bias of InfoDrop}
\label{infodrop_shape_bias}

\begin{figure}[t]
\centering
\centerline{\includegraphics[width=\columnwidth]{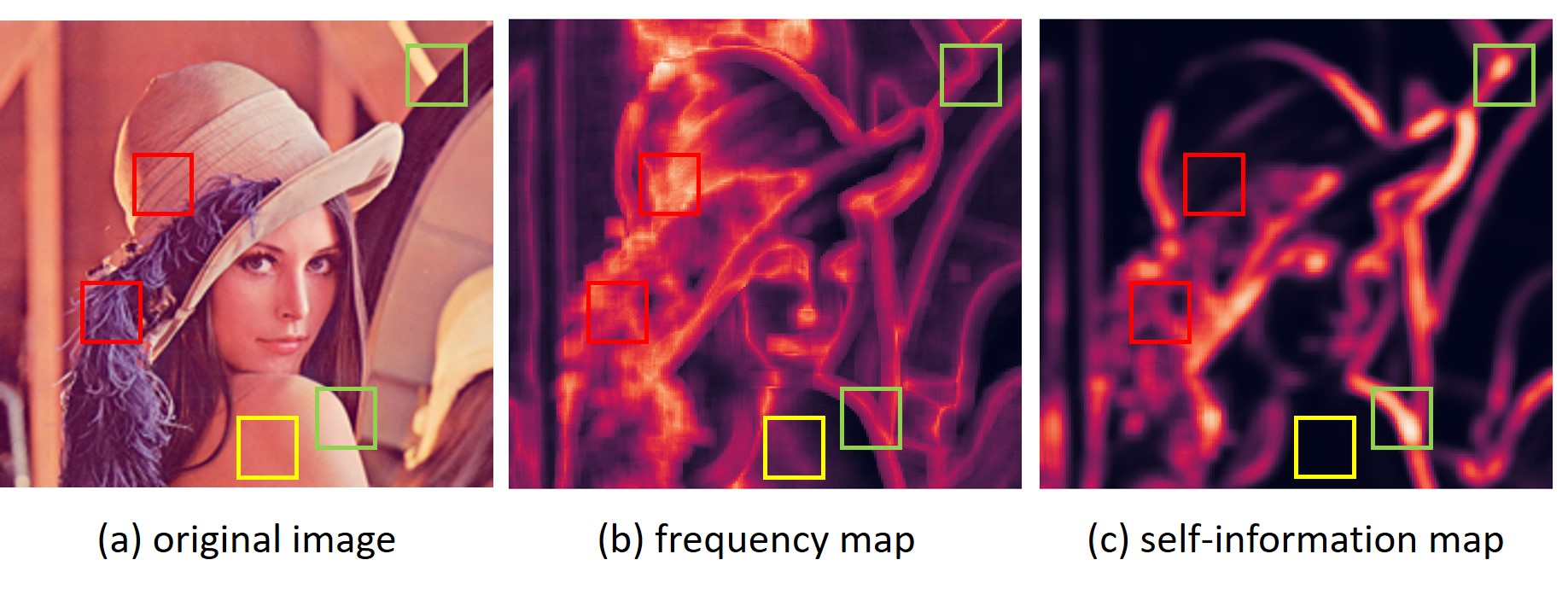}}
\vskip -0.01in
\caption{Picture of Lenna, its frequency map and self-information map. Lighter regions indicate higher values.}
\label{freq_vs_info}

\end{figure}

We conduct several experiments, both qualitatively and quantitatively, to analyze the shape-bias property of InfoDrop. Due to limited space, we refer readers to Appendix for more visualization and detailed experimental settings.

\textbf{A Frequency Perspective} \ \ We first analyze the shape-bias property of self-information by visualizing how it responds to local regions with different spatial frequency. To obtain the average frequency of a local region, we apply Discrete Cosine Transform (DCT)~\cite{ahmed1974discrete} to the local $8\times8$ patch to get the power spectrum, which is further used as weights of each frequency level to get the average frequency. We repeat the process for each position and get the frequency map (Fig.~\ref{freq_vs_info}(b)). We also calculate each position's self-information (Fig.~\ref{freq_vs_info}(c)). As one can observe, for visual primitives including edges and corners (\textbf{green boxes}), they present medium frequency, but are most highlighted by self-information. High-frequency textures (\textbf{red boxes}), as highlighted in frequency map, however, contain relatively low information due to its high-frequency self-repeating. Flat regions (\textbf{yellow boxes}) are filtered by both frequency and information map. This is also consistent with our previous discussions.

\begin{figure}[t]

\centering
\centerline{\includegraphics[width=\columnwidth]{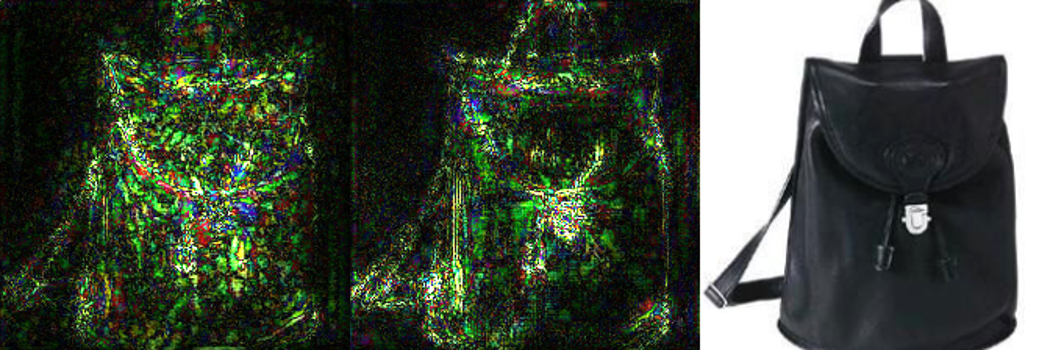}}
\caption{Gradient-based saliency maps of regular CNN (left) and CNN with InfoDrop (middle). Input image is shown on the right.}
\label{grad_map}

\end{figure}

\textbf{Saliency Map of CNN} \ \ To verify the shape-bias InfoDrop brings to CNNs, we visualize gradients of model output \wrt input pixels, which serve as a ``saliency map'' of the network. Specifically we use SmoothGrad~\cite{smilkov2017smoothgrad} to calculate saliency map $\textit{S}(x)$,
\begin{equation}
    \textit{S}(x) = \frac{1}{n}\sum_{i = 1}^{n}\frac{\partial \textit{f}(x_i)}{\partial x_i},
\end{equation}
where $x_i = x + \delta_i$ is original image $x$ with i.i.d. Gaussian noise $\delta_i$, and $\textit{f}(\cdot)$ is the network. An example is shown in Fig.~\ref{grad_map}. We can see that InfoDrop is more human-aligned,  sensitive to shapes of objects, while the saliency map of regular CNN is more noisy and less shape-biased, lacking interpretability.

\begin{table}[t]
\caption{Degradation of classification accuracy on patch-shuffled images. Each image is divided into $m \times m$ patches. Here we use $m = 1$ as baseline, referring to accuracy on original images.}
\label{result_shuffle}
\vskip 0.15in
\begin{center}
\begin{small}
\begin{sc}
\begin{tabular}{c | cccc}
\toprule
$m$ & 1 & 2 & 3 & 4 \\
\midrule
Regular CNN & 99.88 & 99.16 & 97.60 & 92.99 \\
w/ InfoDrop & 99.80 & 95.37 & 89.03 & 79.90 \\
\bottomrule
\end{tabular}
\end{sc}
\end{small}
\end{center}
\end{table}

\textbf{Patch Shuffling} \ \ We also evaluate the shape-bias of InfoDrop through recognizing images whose shape information is ruined but texture is retained. Following~\cite{zhang2019interpreting}, we achieve this goal by dividing images into $m \times m$ patches and randomly shuffling them. Through patch shuffling, global structure is ruined while local texture in each patch is left untouched. We train our model on clean images and test on patch-shuffled test set. We set different values of $m$ and results are listed in Table~\ref{result_shuffle}. Note that $m=1$ means no shuffling is used. As $m$ goes up, global structures are severely ruined, causing a rapid declination in InfoDrop's performance. However, regular CNN is barely influenced since most texture information is preserved. This also indicates that CNN with InfoDrop is more biased towards shape information.

\begin{figure}[t]
\centering
\centerline{\includegraphics[width=\columnwidth]{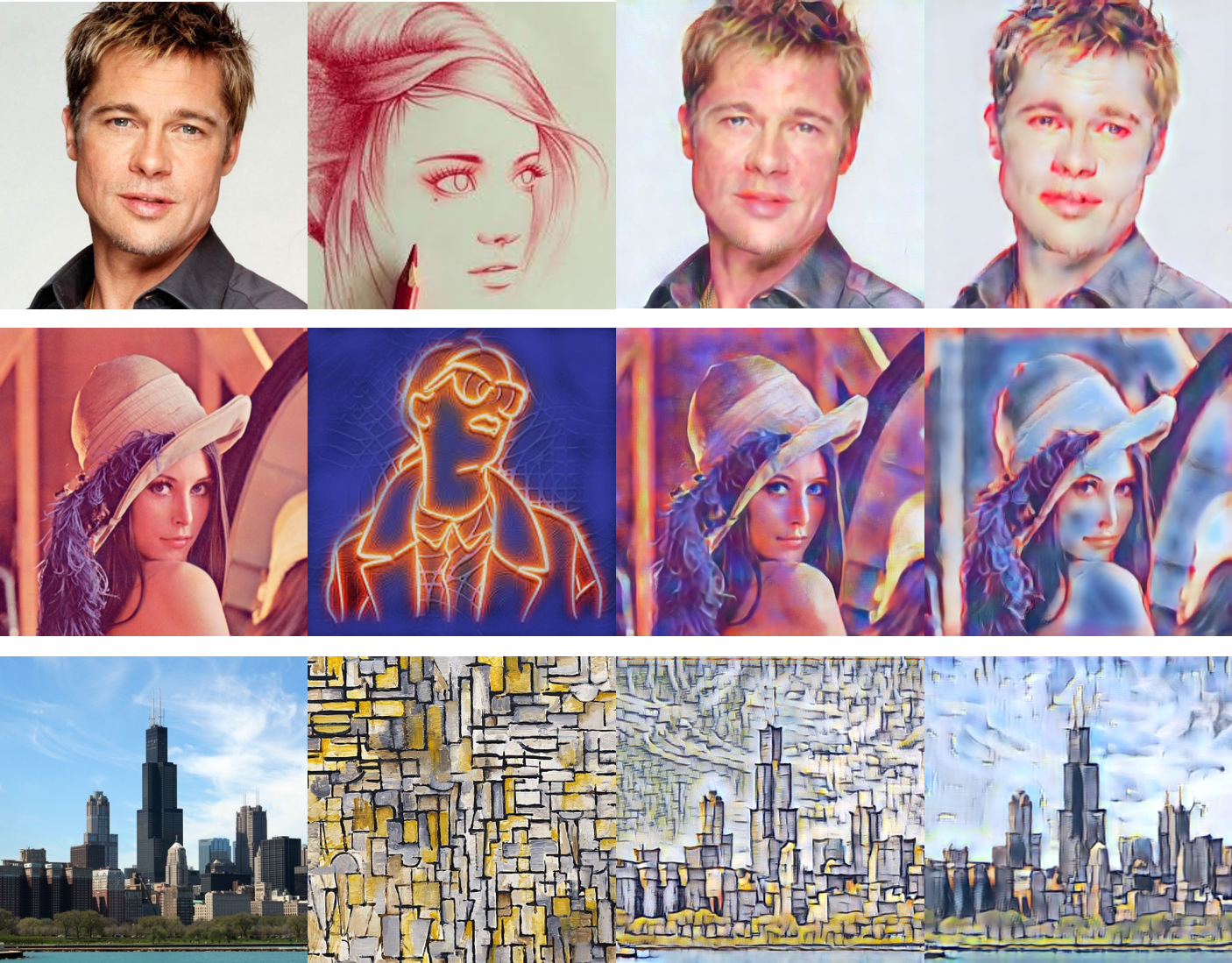}}
\caption{Results of style transfer. From left to right: content image, style image, baseline result, result of InfoDrop. For instance, in baseline result of the last row, both shape (\eg edging) and texture (\eg coloring) style are inherited from style image. However, InfoDrop mainly renders edges in content image, while texture (\eg sky) or color tone is less affected.}
\label{style_transfer}

\end{figure}

\textbf{Style Transfer} \ \ To understand the features extracted by InfoDrop, we conduct ablations in the task of style transfer. Recently,~\citet{huang2017arbitrary} proposed AdaIN algorithm to render a content image with the style of another image (style image). Specifically, features of both content and style images are first extracted by encoder, and then the mean and variance of the content feature is aligned with those of the style feature. Transferred image is then decoded from the aligned content feature. In our experiment, we apply InfoDrop in the encoder and observe changes in the rendered image. By doing so, we expect to see that only the edging style of the content image is rendered by that of the style image, and the texture style is preserved. This is verified by the results in Fig.~\ref{style_transfer}. Take the first row as example, we can see that baseline method mainly change the tone of the whole image. In contrast, InfoDrop inherits the style of red edging and sketching, and applies it on the shape of content image, indicating that InfoDrop is more shape-biased in both content and style images.

\subsection{Robustness of InfoDrop}
\label{infodrop_robustness}

In this section, we first evaluate various kinds of robustness (against distribution shift, image corruption and adversarial perturbation) of InfoDrop through four different tasks (Sec.~\ref{sec:dg} $\sim$ Sec.~\ref{sec:adversary}). Since InfoDrop can be applied to any CNN-based models, and extensive exploration of more complicated base models is beyond the main scope of our studies in this section, we only use simple architecture (\eg ResNet~\cite{he2016deep}) and baseline algorithms, and observe incremental results when InfoDrop is applied. Then we compare InfoDrop with other approaches towards shape-bias (Sec.~\ref{sec:shape_bias_comparison}). Due to limited space, detailed
experimental configuration and additional results are deferred to Appendix. 

\subsubsection{Domain Generalization}
\label{sec:dg}

Due to the natural data variance induced by time, location, weather, \etc, it's a significant feature for visual models to generalize across different domains. To this end, the task of \emph{domain adaptation} is proposed, where labeled data from source domain and unlabeled data from target domain are provided~\cite{shimodaira2000improving}. Prior arts mainly focus on diminishing the distribution shift in feature space between source and target domain~\cite{gretton2007kernel,gretton2009covariate,long2015learning}. A more challenging task, namely \emph{domain generalization}, is later proposed, where data from target domain is unavailable during training. Previous solutions include learning invariant features~\cite{muandet2013domain}, or utilizing auxiliary tasks~\cite{carlucci2019domain}.

In our experiment, we use the naive algorithm as baseline: training a classification model on source domain, and testing on target domain. Following the literature~\cite{carlucci2019domain}, we use PACS~\cite{li2017deeper} as dataset, which consists of four domains, \viz photo, art, cartoon and sketch.

\begin{table}[t]
\caption{Incremental results of single-source domain generalization. + (-) indicates performance gain (decline) from InfoDrop. }
\label{result_single_source_dg}
\vskip 0.15in
\centering
\begin{small}
\begin{sc}
\begin{tabular}{c  cccc}
\toprule
\diagbox[width=8em]{Source}{Target} & photo & art & cartoon & sketch \\
\midrule
photo &   \textit{-0.06}    & \textbf{+2.49} & \textbf{+6.52} & \textbf{+6.09} \\
art   & \textbf{+0.12} &  \textit{+0.20}    & \textbf{+1.57} & \textbf{+0.81}  \\
cartoon & -0.84 & -0.44 &  \textit{+0.04}    & \textbf{+4.81} \\
sketch & \textbf{+11.91} & \textbf{+4.23} & \textbf{+6.19} &   \textit{+0.15}    \\
\bottomrule
\end{tabular}
\end{sc}
\end{small}

\vskip -0.01in
\end{table}

Results on single-source domain generalization are shown in Table~\ref{result_single_source_dg}. Here we report the relative improvement of InfoDrop over baseline. For the absolute accuracies, please refer to Appendix. Compared with baseline, InfoDrop boosts performances in multiple settings, especially with sketch as the source or target domain. This also reflects the shape-bias of InfoDrop, considering that sketches mainly consist of shape information. It is also worth noticing that our model can keep the performance on the \emph{source} domain after InfoDrop is applied.

\begin{table}[t]
\caption{Results on multi-source domain generalization. Performance of JiGen~\cite{carlucci2019domain} and D-SAM~\cite{d2018domain} are listed for comparison. }
\label{result_multi_dource_dg}
\vskip 0.15in
\centering
\begin{small}
\begin{sc}
\begin{tabular}{c | cccc}
\toprule
\diagbox[width=8em]{Methods}{Target} & photo & art & cartoon & sketch \\
\midrule
D-SAM & 95.30 & 77.33 & 72.43 & \textbf{77.83} \\
JiGen   & 96.03 & 79.42 & 75.25 & 71.35  \\
Baseline & 95.98 & 77.87 & 74.86 & 70.17 \\
+ InfoDrop & \textbf{96.11} & \textbf{80.27} & \textbf{76.54} & 76.38  \\
\bottomrule
\end{tabular}
\end{sc}
\end{small}

\end{table}

We also obtain results on multi-source domain generalization.  Table~\ref{result_multi_dource_dg} shows results on each domain after trained on other three domains. When trained with InfoDrop, the model is more robust to the distribution shift between different domains, and obtains consistent improvements over all target domains. Moreover, the vanilla baseline with InfoDrop is already better than or comparable with other state-of-the-art methods on each target domain.

\subsubsection{Few-shot Classification}
\label{sec:few-shot}

\begin{table*}[ht]
\caption{Few-shot classification results under different settings with ProtoNet as baseline. All experiments are 5-way. Usage of data augmentation is denoted by `w/', and vice versa.}
\label{result_few_shot_setting}
\vskip 0.15in
\centering
\begin{small}
\begin{sc}
\begin{tabular}{c | cc | cc | cc | cc | cc | cc}
\toprule
 & \multicolumn{4}{c}{CUB} & \multicolumn{4}{c}{\emph{mini}-ImageNet} & \multicolumn{4}{c}{\emph{mini}-ImageNet$\rightarrow$CUB} \\
 & \multicolumn{2}{c}{5-shot} & \multicolumn{2}{c}{1-shot} & \multicolumn{2}{c}{5-shot} & \multicolumn{2}{c}{1-shot} & \multicolumn{2}{c}{5-shot} & \multicolumn{2}{c}{1-shot} \\
 &  w/o  & w/  & w/o  & w/  & w/o  & w/  & w/o  & w/  & w/o  & w/  & w/o & w/ \\
\midrule
ProtoNet & 67.13 & 77.64 & 51.62 & 58.83 & 63.84 & 66.85 & 47.96 & 47.17 & 52.71 & 54.62 & \textbf{39.36} & 35.24 \\
+ InfoDrop & \textbf{70.94} & \textbf{78.18} & \textbf{52.40} & \textbf{59.06} & \textbf{66.85} & \textbf{67.25} & \textbf{49.61} & \textbf{50.09} & \textbf{55.06} & \textbf{55.09} & 37.11 & \textbf{37.50} \\
\bottomrule
\end{tabular}
\end{sc}
\end{small}

\vskip 0.0in
\end{table*}

Current CNNs rely on huge amount of labeled data to learn powerful representations for downstream tasks. However, the learned representations may generalize poorly to unseen objects and scenes. This is in contrast to the human visual system, which is able to quickly grasp the feature of an unseen object given only a few examples. To this end, the task of \emph{few-shot classification} is proposed, where a model needs to recognize classes unseen during training with limited examples. The main challenge here is the huge class-wise distribution shift. Following the literature, we use `$m$-way $n$-shot classification' to refer to the setting where test data come from $m$ novel classes each with $n$ examples provided.

Following the setting in~\citet{chen2018a}, we evaluate InfoDrop on two popular datasets: CUB~\cite{wah2011caltech} and \emph{mini}-ImageNet~\cite{ravi2016optimization}, meanwhile also test our model in the cross-domain scenario~\cite{chen2018a}, where \emph{mini}-ImageNet is used for training and CUB for testing. We denote this setting by \emph{mini}-ImageNet$\rightarrow$CUB. For a full comparison, we test models trained both with and without data augmentation. For baseline algorithms, we follow~\citet{chen2018a} and adopt three common approaches, \viz ProtoNet~\cite{snell2017prototypical}, MatchingNet~\cite{vinyals2016matching} and RelationNet~\cite{sung2018learning}.

\begin{table}[t]
\caption{Few-shot classification results with different baseline methods. All results are from 5-way classification on CUB without data augmentation.}
\label{result_few_shot_method}
\vskip 0.15in
\centering
\begin{small}
\begin{sc}
\begin{tabular}{c | cc}
\toprule
 & 5-shot & 1-shot \\
\midrule \midrule
MatchingNet & 71.18 $\pm$ 0.70 & 57.81 $\pm$ 0.88 \\
+ InfoDrop & \textbf{72.32 $\pm$ 0.69} & \textbf{57.88 $\pm$ 0.91} \\
\midrule
ProtoNet & 67.13 $\pm$ 0.74 & 51.62 $\pm$ 0.90 \\
+ InfoDrop & \textbf{70.94 $\pm$ 0.72} & \textbf{52.40 $\pm$ 0.90} \\
\midrule
RelationNet & 69.85 $\pm$ 0.75 & 56.71 $\pm$ 1.01 \\
+ InfoDrop & \textbf{73.72 $\pm$ 0.71} & \textbf{59.21 $\pm$ 0.98} \\
\bottomrule
\end{tabular}
\end{sc}
\end{small}

\vskip 0.05in
\end{table}

First we use ProtoNet as baseline and evaluate our method under different settings (Table~\ref{result_few_shot_setting}). Under almost all the settings, InfoDrop brings a non-trivial improvement in performance. One may notice that improvements on \emph{mini}-ImageNet are larger than CUB, which is reasonable due to the larger distribution shift to overcome in \emph{mini}-ImageNet~\cite{chen2018a}. As another observation, the improvements on 5-shot classification is larger than 1-shot. This implies that despite the robustness of shape features, they may not be as discriminative as texture features, hence requiring more examples for recognition. As a consequence, we may still need \emph{some} texture to learn a discriminative and robust model (Sec.~\ref{sec:is_shape_all_we_need}).  Also, note that for baseline method, sometimes data augmentation may damage performance, which is possibly because augmentation leads to overfitting in the base classes. However, similar behavior is not observed on InfoDrop.

Then we check whether InfoDrop can bring a consistent improvement on different baselines. As shown in Table~\ref{result_few_shot_method}, on three baseline methods, InfoDrop improves the robustness universally. Note that InfoDrop most benefits RelationNet, possibly because its relation head learns a better similarity metrics between complex shapes.

\subsubsection{Robustness against Image Corruption}
\label{sec:corruption}

\begin{table*}[t]
\caption{Classification accuracy on clean and randomly corrupted images. `A' and `I' means usage of adversarial training and InfoDrop, respectively. All corruptions are generated under severity of level 1~\cite{hendrycks2019benchmarking}.}
\label{result_corruption}
\vskip 0.15in
\centering
\begin{footnotesize}
\begin{sc}
\setlength\tabcolsep{2.0pt}
\begin{tabular}{cc | c | ccc | ccc | ccc | ccc}
\toprule
A & I & clean & \multicolumn{3}{c}{noise} & \multicolumn{3}{c}{blur} & \multicolumn{3}{c}{weather} & \multicolumn{3}{c}{digital} \\
 & & & gaussian & shot & impulse & defocus & motion & gaussian & snow & frost & fog & elastic & jpeg & saturate \\
\midrule
\xmark & \xmark & 82.98 & 66.38 & 62.85 & 49.97 & \textbf{65.97} & \textbf{74.79} & \textbf{78.75} & 53.10 & 67.09 & 72.42 & \textbf{76.58} & 79.77 & 77.15 \\
\xmark & \cmark & \textbf{83.14} & 69.58 & 66.83 & 53.00 & 62.52 & 71.76 & 77.03 & 56.44 & \textbf{69.80} & \textbf{72.75} & 74.54 & \textbf{80.49} & \textbf{77.77} \\
\cmark & \xmark & 79.69 & 75.30 & 73.80 & 70.71 & 61.53 & 71.68 & 73.77 & 61.11 & 69.06 & 54.52 & 71.69 & 79.31 & 72.62\\
\cmark & \cmark & 78.59 & \textbf{76.17} & \textbf{74.90} & \textbf{72.26} & 62.32 & 71.32 & 74.04 & \textbf{61.69} & \textbf{69.83} & 55.00 & 70.26 & 78.10 & 71.26 \\
\bottomrule
\end{tabular}
\end{sc}
\end{footnotesize}

\vskip -0.1in
\end{table*}



It is essential for visual models to give stable predictions under various kinds of corruptions (\eg weather, blur, noise), especially in safety-critical situations. However, current CNNs are vulnerable to random corruptions and hardly generalize to different kinds of corruptions when trained on a specific one~\cite{dodge2017study}. Recently, \citet{geirhos2018imagenet} find that a \emph{consistently} improved robustness against different corruptions can be achieved by training a shape-biased model. In~\citet{hendrycks2019benchmarking}, benchmarks of model robustness are provided on 18 common types of corruption. In our experiments, we apply the same corruption functions on Caltech-256 dataset~\cite{griffin2007caltech} to test the robustness of InfoDrop. For comparison, we also test robustness of adversarially trained networks with and without InfoDrop. Adversarial training is known to improve robustness to noise and blur corruptions, while degrade performance on some others (\eg fog, contrast)~\cite{gilmer2019adversarial}. Results are shown in Table~\ref{result_corruption}. Due to limited space, we only show 12 types of corruptions here. Full comparisons can be found in Appendix. Clearly, InfoDrop improves baseline's robustness against most corruptions (\eg noise, weather, digital) universally, although no noisy data is used for training.
This also implies the potential of InfoDrop to generalize to other untested types of corruptions. Nonetheless, the performance may further degrade under blurring nonetheless, which is reasonable because blurring brings more distortion of shapes while others mainly corrupts texture information. It is also noticeable that InfoDrop can be incorporated with adversarial training and obtain even better robustness with little overhead.

\subsubsection{Adversarial Robustness}
\label{sec:adversary}



\begin{table}[t]
\caption{Adversarial robustness under different perturbation norm on CIFAR-10. `A' and `I' refer to the usage of adversarial training and InfoDrop, respectively.}
\label{result_adv}
\vskip 0.15in
\centering
\begin{small}
\begin{sc}
\begin{tabular}{cc | cccc}
\toprule
A & I & $\ell_\infty = 0$ & $\ell_\infty = \frac{1}{255}$ & $\ell_\infty = \frac{2}{255}$ & $\ell_\infty = \frac{8}{255}$ \\
\midrule
\xmark & \xmark & 94.57 & 55.26 & 7.99 & 0.01 \\
\xmark & \cmark & 94.08 & 59.35 & 12.41 & 0.03\\
\cmark & \xmark & 86.62 & 82.03 & 77.44 & 42.05 \\
\cmark & \cmark & 86.50 & 82.06 & 77.41 & 43.19\\
\bottomrule
\end{tabular}
\end{sc}
\end{small}

\end{table}

Except for random corruptions, CNNs are also vulnerable to carefully-designed imperceptible perturbations, namely adversarial perturbations~\cite{szegedy2013intriguing}. This leads to another crucial challenge for current CNN-based models. Most work on adversarial robustness is based on adversarial training~\cite{madry2017towards}. To evaluate adversarial robustness of InfoDrop, we conduct ablations on both baseline and adversarial trained models. Following the literature, we use CIFAR-10~\cite{krizhevsky2009learning}, a widely-reported benchmark. For attacking, we use 20 runs of PGD~\cite{madry2017towards} with constrained $\ell_\infty$ norm in both adversarial training and testing. As shown in Table~\ref{result_adv}, InfoDrop can improve robustness of baseline models under low-norm attack, but it still fails when the perturbation is large. Moreover, InfoDrop can be combined with adversarial training and provide extra robustness. Under the norm $\ell_\infty = \frac{8}{255}$, InfoDrop brings an improvement of 1\% accuracy.

\subsubsection{Comparison with Other Shape-biased Methods}
\label{sec:shape_bias_comparison}

Some approaches have also been proposed recently to train a shape-biased model. For example, \citet{geirhos2018imagenet} propose to train the network on extra images with various texture styles in order to learn the shared shape features. \citet{wang2018learning} propose to use Gray-level Co-occurrence Matrix~\cite{lam1996texture} as an indicator of texture, and decompose the feature from it. Other attempts include using different auxiliary tasks~\cite{wang2019learning,carlucci2019domain}.

\begin{table}[t]
\caption{Performance of different shape-biased methods on single-source domain generalization. Here we use Photo as the source domain, and report the accuracies on the other three target domains. Baseline indicates a simple ResNet50 model. $ ^\dagger$ means extra finetuning on ImageNet is required during pretraining.}
\label{tab:shape_bias_comparison}
\vskip 0.15in
\centering
\begin{small}
\begin{sc}
\begin{tabular}{lccc}
\toprule
 & Art & Cartoon & Sketch \\
\midrule
Baseline & 73.68 & 34.34 & 36.73\\
IN + SIN & 72.80 & 40.04 & 58.70 \\
IN + SIN$^\dagger$ & 74.51 & 38.38 & 42.61  \\
InfoDrop & 74.07 & 41.40 & 54.31 \\
\bottomrule
\end{tabular}
\end{sc}
\end{small}

\end{table}

Here we compare InfoDrop with the approach in~\citet{geirhos2018imagenet}, which pretrains the network on ImageNet (IN) as well as Stylized-ImageNet (SIN). For comparison with other shape-biased methods, please refer to Appendix. Specifically, we evaluate the performances on single-source domain generalization. We compare InfoDrop (pretrained only on IN) with a ResNet50 pretrained on both IN and SIN. Results are shown in Table~\ref{tab:shape_bias_comparison}. We can see that both methods can bring an improvement in the model robustness. Particularly, pretraining on SIN can largely increase the accuracy on Sketch domain, which is probably because SIN already contains images with sketch style. Remarkably, InfoDrop can improve the robustness consistently without seeing any target domain examples beforehand.

\subsection{Ablation Studies}
\label{ablation}

In this section we mainly discuss how different configurations or hyperparameters will impact the performance of InfoDrop. We first start with the role of temperature $T$ in Eq.~\ref{sample_prob}. Intuitively, lower temperature means more conservative filtering, \ie, only patches with the least information (\eg constant-valued regions) are dropped, while most shape and texture are preserved. An infinite temperature, however, will wipe out differences between shape and texture and act in a purely random way as regular Dropout. Apparently, somewhere between is what we intend for, where it can distinguish shape and texture, and filter out the latter. As verification, we conduct ablations on 5-way 1-shot classification on \emph{mini}-ImageNet$\rightarrow$CUB. As shown in Table~\ref{ablation_T}, it reaches the highest accuracy when $T = 1$. Higher or lower $T$ will degrade the performance. This means to be more robust, the model needs to filter out textures whilst preserve shape information, which is consistent with our analysis.

\begin{table}[t]
\caption{Ablation study of temperature $T$ in few-shot classification. Here we use ProtoNet as baseline (denoted by `-'). When $T = inf$, it degrades to regular Dropout.}
\label{ablation_T}
\vskip 0.15in
\centering
\begin{small}
\begin{sc}
\begin{tabular}{ccccccc}
\toprule
$T$ & - & 0.1 & 0.5 & 1.0 & 3.0 & inf \\
\midrule
Acc & 35.24 & 36.33 & 37.50 & 37.89 & 36.21 & 35.54 \\
\bottomrule
\end{tabular}
\end{sc}
\end{small}

\end{table}

\begin{table}[t]
\caption{Ablation study on number of residual blocks equipped with InfoDrop. Results show the accuracies in domain generalization from sketch to art.}
\label{ablation_layer}
\vskip 0.15in
\centering
\begin{small}
\begin{sc}
\begin{tabular}{c | ccccc}
\toprule
Blocks & 0 & $0^+$ & 1 & 2 & 3 \\
\midrule
Accuracy & 27.34 & 29.12 & 31.05 & 28.85 & 29.28 \\
\bottomrule
\end{tabular}
\end{sc}
\end{small}
\end{table}

Now we discuss to which layers should InfoDrop be applied. Technically, it can be integrated into any convolutional layers. But since InfoDrop extracts local self-information and locate important primitives, intuitively, as a local algorithm, it should be applied to lower layers of a CNN. In our experiments, we apply InfoDrop to the first $K$ residual blocks of ResNet18, where $K = 0, 0^+, 1, 2, 3$, where $0^+$ means InfoDrop is applied only to the first convolutional layer before all residual blocks. As shown in Table~\ref{ablation_layer}, $K=1$ gives the best performance. For higher layers, extracted features are more abstract and dropping them may degrade performance.

\subsection{Is Shape Information All You Need?}
\label{sec:is_shape_all_we_need}

\begin{figure}[t]
\centering
\centerline{\includegraphics[width=0.9\columnwidth]{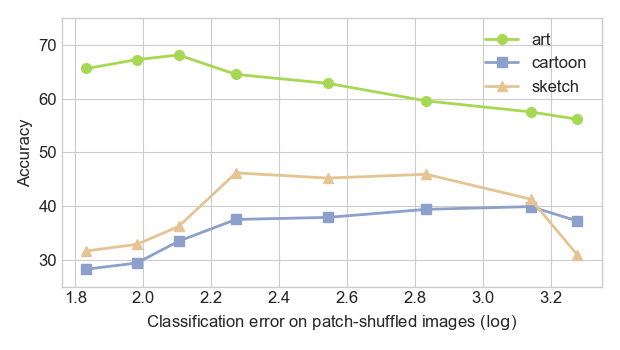}}
\caption{Domain generalization performance of models with different levels of shape-bias. The x-axis is the classification error on images with shuffled (3$\times$3) patches, which is used as a indicator of the shape-bias level, \ie, models with larger shape-bias tend to fail to recognize patch-shuffled images.}
\label{fig:is_shape_all_we_need}

\end{figure}

In previous sections we have demonstrated how shape-bias can benefit CNN's robustness under different scenarios. This raises another question: how biased should our model be? For example, does a visual model still work well if it only perceives shape information? The answer may be ``no'', considering that texture information plays a different but also important role in the human visual system (\eg multi-modal perception~\cite{sann2007perception}). It is also verified in experiments on deep models~\cite{xiao2019shape} that shape itself does not suffice for high-quality visual recognition. Intuitively, there should exist an optimal ``bias level'' so that the model can be robust enough and meanwhile recognize objects with a proper precision, and this optimal level may vary from task to task.

To verify this, we conduct experiments on domain generalization. Specifically, we tune the temperature $T$ to train models with different levels of shape-bias. To quantify the shape-bias, we use the classification error on patch-shuffled images as an indicator, considering that larger shape-bias generally leads to higher classification error on patch-shuffled images. We use photo as source domain, and test the performances on art, cartoon and sketch. As shown in Fig.~\ref{fig:is_shape_all_we_need}, the performances on all target domains all go through an ascending at first, and then fall back when the shape-bias keeps being enhanced. Moreover, different target domains prefer different optimal bias levels.
This implies that current CNNs are overly texture-biased, and we need to reach a ``sweet spot'' between shape and texture.\footnote{Another question would be what is the proper relationship between shape and texture? Should they act like two separate cues in a parallel way, or in a hierarchical way, where shape first provides a quick, coarse recognition, and then details are observed through texture? We leave this for further exploration.}





\section{Conclusion}

In this work, we aim at universally improving various kinds of robustness of CNN by alleviating its texture-bias. To reduce texture-bias, we get our inspiration from the human visual system and propose Informative Dropout, an effective model-agnostic algorithm. We detect texture and shape by the local self-information in an image, and use a Dropout-like algorithm to decorrelate the model output from the local texture. Through extensive experiments we observe improved shape-bias as well as various kinds of robustness. Furthermore, we find our method can be incorporated with other algorithms (\eg adversarial training) and achieve higher robustness. Through this work, we shed some light on the relationship between CNN's shape-bias and  robustness, as well as new approaches to trustworthy machine learning algorithms.

\begin{small}
\section*{Acknowledgement}
Prof. Yadong Mu is partly supported by National Key R\&D Program of China (2018AAA0100702) and Beijing Natural Science Foundation (Z190001).
Dr. Zhanxing Zhu is supported by National Natural Science Foundation  of  China  (No.61806009 and 61932001),  PKU-Baidu Funding 2019BD005 and Beijing Academy of Artificial  Intelligence  (BAAI).
Dinghuai Zhang is supported by the Elite Undergraduate Training Program of Applied Math of the School of Mathematical Sciences at Peking University. The authors are thankful to Tianyuan Zhang, Dejia Xu, Yiwen Guo and the anonymous reviewers for the insightful discussions and useful suggestions.
\end{small}



\bibliography{references}
\bibliographystyle{icml2020}


\appendix

\begin{table*}[h!]
\caption{Comparison between InfoDrop and other state-of-the-art results of multi-source domain generalization on PACS dataset. We use $^\star$ to denote the setting in~\citet{carlucci2019domain} (\eg extra data augmentation, different train-test split, and different learning rate scheduling). Reported state-of-the-art methods include DSN~\cite{bousmalis2016domain}, LCNN~\cite{li2017deeper}, MLDG~\cite{li2018learning}, Fusion~\cite{mancini2018best}, MetaReg~\cite{balaji2018metareg}, JiGen~\cite{carlucci2019domain}, HEX~\cite{wang2018learning} and PAR~\cite{wang2019learning}. We use PAR$_B$, PAR$_M$, PAR$_H$ to denote PAR with broader local pattern, more powerful pattern classifier and higher level of local concept, respectively. For more details, please refer to the original paper~\cite{wang2019learning}.}
\label{dg_alexnet}
\vskip 0.1in
\centering
\begin{sc}
\begin{tabular}{c c c| c c c c c}
\toprule
 & Domain ID & Data Aug. & Art & Cartoon & Photo & Sketch & Average \\
\midrule \midrule
AlexNet & \xmark & \xmark & 63.3 & 63.1 & 87.7 & 54 &  67.03  \\
\midrule
DSN & \cmark & \xmark  & 61.1 & 66.5 & 83.2 & 58.5 & 67.33 \\
L-CNN & \cmark & \xmark & 62.8 & 66.9 & 89.5 & 57.5 & 69.18 \\
MLDG & \cmark & \xmark & 63.6 & 63.4 & 87.8 & 54.9 & 67.43 \\
Fusion & \cmark & \xmark & 64.1 & 66.8 & 90.2 & \textbf{60.1} & 70.30 \\
MetaReg & \cmark & \xmark & \textbf{69.8} & \textbf{70.4} & \textbf{91.1} & 59.2 & \textbf{72.63} \\
\midrule
HEX & \xmark & \xmark & 66.8 & 69.7 & 87.9 & 56.3 & 70.18 \\
PAR & \xmark & \xmark & \textbf{66.9} & 67.1 & 88.6 & 62.6 & 71.30 \\
PAR$_B$ & \xmark & \xmark & 66.3 & 67.8 & 87.2 & 61.8 & 70.78 \\
PAR$_M$ & \xmark & \xmark & 65.7 & 68.1 & 88.9 & 61.7 & 71.10 \\
PAR$_H$ & \xmark & \xmark & 66.3 & \textbf{68.3} & \textbf{89.6} & \textbf{64.1} & \textbf{72.08} \\
\midrule
JiGen$^\star$ & \xmark & \cmark & 67.6 & \textbf{71.7} & 89.0 & 65.1 & 73.38 \\
PAR$^\star$ & \xmark & \cmark & 68.0 & 71.6 & \textbf{90.8} & 61.8 & 73.05 \\
PAR$_B$$^\star$ & \xmark & \cmark & 67.6 & 70.7 & 90.1 & 62.0 & 72.59 \\
PAR$_M$$^\star$ & \xmark & \cmark & 68.7 & 71.5 & 90.5 & 62.6 & 73.33  \\
PAR$_H$$^\star$ & \xmark & \cmark & 68.7 & 70.5 & 90.4 & 64.6 & 73.54 \\
InfoDrop$^\star$ & \xmark & \cmark & \textbf{70.3} & \textbf{71.7} & 90.3 & \textbf{70.6} & \textit{\textbf{75.73}} \\
\bottomrule
\end{tabular}
\end{sc}
\end{table*}

\begin{table*}[h!]
\caption{Robustness against 18 types of image corruptions. With vanilla CNN or adversarially trained CNN as baseline, InfoDrop can improve robustness against most corruptions consistently.}
\label{result_corruption}
\vskip 0.15in
\centering
\begin{sc}
\begin{tabular}{l c c c c}
\toprule
Corruption type & Vanilla & + InfoDrop & + Adv. Train & + Info \& Adv \\
\midrule
Gaussian noise & 66.38 & 69.58 & 75.30 & \textbf{76.17} \\
Shot noise & 62.85 & 66.83 & 73.80 & \textbf{74.90} \\
Impulse noise & 49.97 & 53.00 & 70.71 & \textbf{72.26} \\
Defocus blur & \textbf{65.97} & 62.52 & 61.53 & 62.32 \\
Motion Blur & \textbf{74.79} & 71.76 & 71.68 & 71.32 \\
Zoom blur & \textbf{62.92} & 58.56 & 61.58 & 60.29 \\
Snow & 53.10 & 56.44 & 61.11 & \textbf{61.69} \\
Frost & 67.09 & \textbf{69.80} & 69.06 & \textbf{69.83} \\
Fog & 72.42 & \textbf{72.75} & 54.52 & 55.00 \\
Brightness & 82.20 & \textbf{82.72} & 79.08 & 78.33 \\
Contrast & \textbf{76.66} & 75.07 & 57.93 & 57.96 \\
Elastic Transform & \textbf{76.58} & 74.54 & 71.69 & 70.26 \\
Pixelate & 79.53 & \textbf{79.81} & 78.51 & 77.66 \\
Jpeg compression & 79.77 & \textbf{80.49} & 79.31 & 78.10 \\
Speckle noise & 66.19 & 69.54 & 74.74 & \textbf{75.66} \\
Gaussian blur & \textbf{78.75} & 77.03 & 73.77 & 74.04 \\
Spatter & 79.18 & \textbf{79.66} & 78.04 & 75.55 \\
Saturate & 77.15 & \textbf{77.77} & 72.62 & 71.26\\
\bottomrule
\end{tabular}
\end{sc}
\end{table*}

\begin{table*}[ht]
\vskip -0.1in
\caption{InfoDrop's improvement in absolute accuracies on single-source domain generalization.}
\label{tab:single_source_dg_abs}
\vskip 0.15in
\centering
\begin{small}
\begin{sc}
\begin{tabular}{c  cccc}
\toprule
\diagbox[width=8em]{Source}{Target} & photo & art & cartoon & sketch \\
\midrule
photo &   \textit{99.88 $\rightarrow$ 99.82}    & \textbf{66.21 $\rightarrow$ 68.70} & \textbf{24.15 $\rightarrow$ 30.67} & \textbf{33.60 $\rightarrow$ 48.36} \\
art   & \textbf{96.71 $\rightarrow$ 96.83} &  \textit{96.46 $\rightarrow$ 96.66}    & \textbf{59.77 $\rightarrow$ 61.22} & \textbf{56.35 $\rightarrow$ 57.16}  \\
cartoon & 86.41 $\rightarrow$ 85.57 & 69.29 $\rightarrow$ 68.85 &  \textit{99.53 $\rightarrow$ 99.57}    & \textbf{64.85 $\rightarrow$ 69.66} \\
sketch & \textbf{32.34 $\rightarrow$ 44.25} & \textbf{27.34 $\rightarrow$ 31.57} & \textbf{43.81 $\rightarrow$ 50.00} &   \textit{99.47 $\rightarrow$ 99.62}    \\
\bottomrule
\end{tabular}
\end{sc}
\end{small}

\end{table*}

\newpage

\section{Additional Results}

\subsection{Results on Domain Generalization and Comparison with Other Shape-biased Models}

Several shape-biased methods have recently been proposed to learn robust representations under different domains~\cite{carlucci2019domain,wang2019learning,wang2018learning}. For a full comparison with these state-of-the-art methods, we also test performance of InfoDrop on domain generalization with AlexNet~\cite{krizhevsky2012imagenet} as backbone. We follow the setting in~\citet{wang2019learning}. Results are shown in Table~\ref{dg_alexnet}. On all four domains, InfoDrop with vanilla AlexNet as baseline is already better than or comparable to other state-of-the-art methods. Note that among these methods, JiGen, HEX and PAR are methods which explicitly train a shape-biased model.

The absolute accuracies of single-source domain generalization are also postponed here (Table~\ref{tab:single_source_dg_abs}) due to the limited space in Table~\ref{result_single_source_dg}.



\subsection{Robustness Against Image Corruption}

Complete results of robustness against image corruption are shown in Table~\ref{result_corruption}. As baseline, we use both vanilla CNN and adversarially trained CNN. Then we apply InfoDrop and report the improved results. As shown in the table, on both baselines InfoDrop can improve robustness against most corruptions non-trivially.

\subsection{Visualization of Saliency Map}

\begin{figure*}[hp]
\vskip 0.2in
\centering
\centerline{\includegraphics[width=\textwidth]{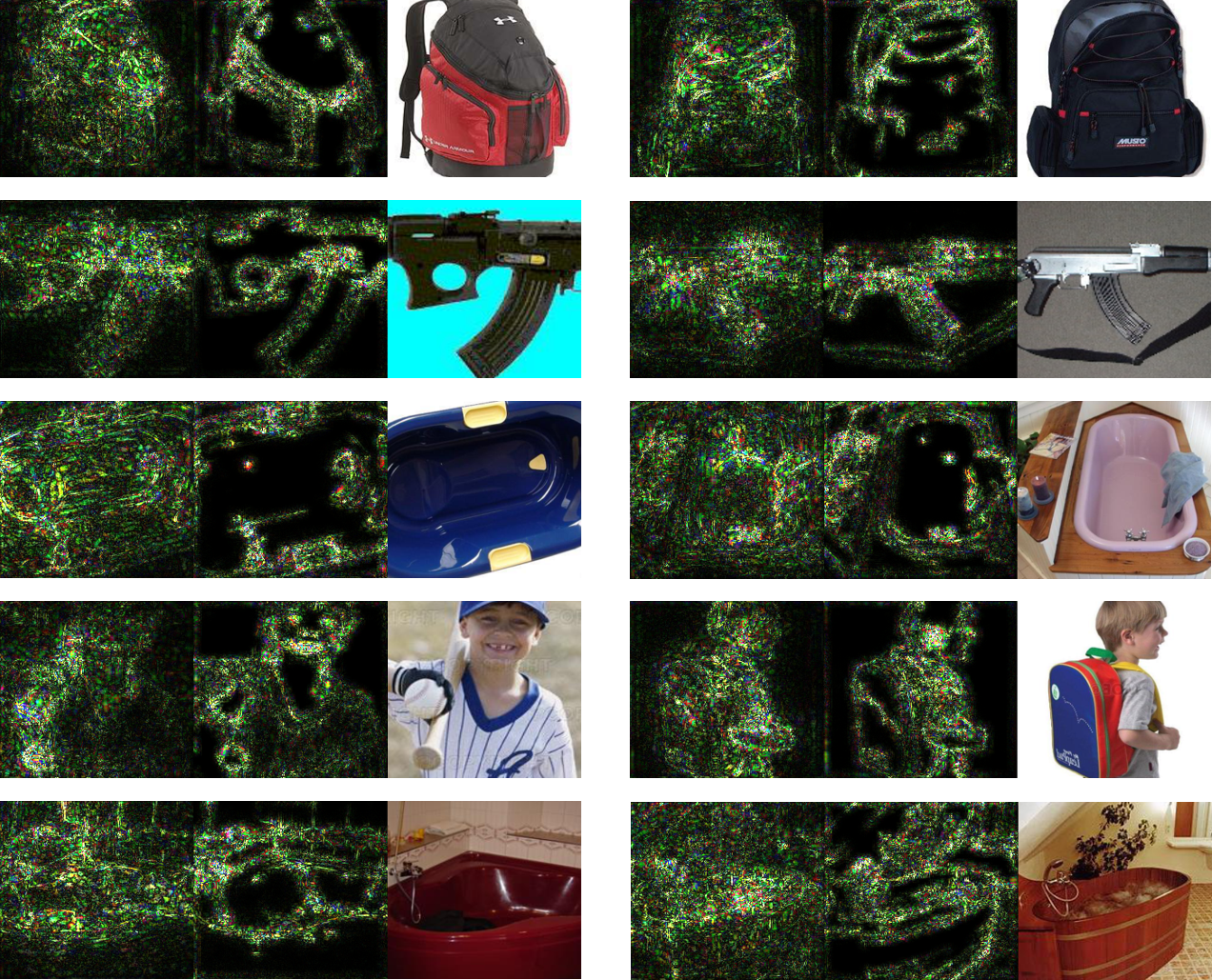}}
\caption{Visualization of CNN's saliency (gradient) map. For each subfigure, from left to right: saliency map of vanilla CNN, saliency map of CNN with InfoDrop and original image.}
\label{supp_grad}
\vskip -0.2in
\end{figure*}

\begin{figure*}[hp!]
\vskip 0.2in
\centering
\centerline{\includegraphics[width=\textwidth]{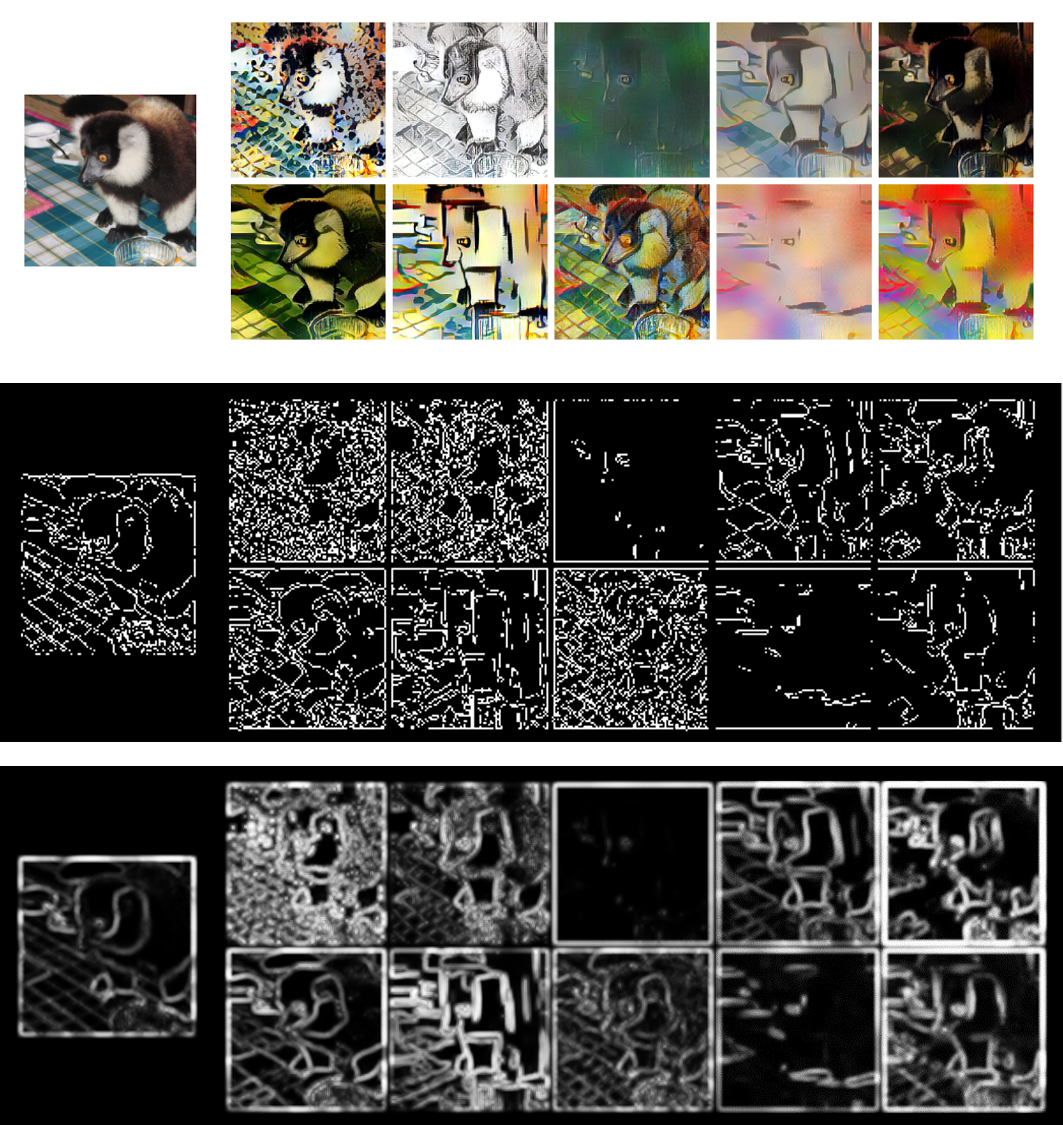}}
\caption{\emph{Top}: The original image and 10 different stylized version. Stylized images in the example can be found in~https://github.com/rgeirhos/Stylized-ImageNet. \emph{Middle}: Edge detecting results of stylized images. \emph{Bottom}: Distribution of self-information in each image.}
\label{supp_info}
\vskip -0.2in
\end{figure*}

Additional visualization results of saliency map of InfoDrop are plotted in Fig.~\ref{supp_grad}. For comparison, saliency map of vanilla CNN is also displayed. Obviously, saliency of InfoDrop is more biased towards global structure, thus more human-aligned and interpretable.

\subsection{Visualization of Self-Information}

We further visualize self-information on the dataset of Stylized-Imagenet~\cite{geirhos2018imagenet}. As a comparison, we also show the results of edge detecting. As shown in Fig.~\ref{supp_info} (Top), the original image is stylized with different art work. As a result (Middle), edge detecting is largely influenced by texture information in different style, sometimes even ruining the image content severely. However, distribution of self-information in each stylized image keeps mostly the same, accentuating global structure and meanwhile repressing local texture.


\section{Experimental Settings}

Of all hyper-parameters, we find $r_0$ and $T$ the most important for model performance. For all the tasks, we search $r_0$ in $[0.1, 2.0]$ and $T$ in $[0.01, 1.0]$. We fix $h = 1$, $R = 3$ through the whole experiment. For ResNet18~\cite{he2016deep}, we apply InfoDrop in both the first convolutional layer and the first residual block, or just in the first layer under some settings. All hyper-parameters are selected according to results on validation set. We use PyTorch~\cite{paszke2019pytorch} for implementation and train all the models on single NVIDIA Tesla P100 GPU.

\subsection{Domain Generalization}

\subsubsection{Dataset}

We use PACS~\cite{li2017deeper} as our dataset for domain generalization. PACS consists of four domains (photo, art painting, cartoon and sketch), each containing 7 categories (dog, elephant, giraffe, guitar, horse, house and person). The dataset is created by intersecting classes in Caltech-256~\cite{griffin2007caltech}, Sketchy~\cite{sangkloy2016sketchy}, TU-Berlin~\cite{eitz2012humans} and Google Images. Dataset can be downloaded from http://sketchx.eecs.qmul.ac.uk/. Following protocol in~\citet{li2017deeper}, we split the images from training domains to 9 (train) : 1 (val) and test on the whole target domain. We use a simple data augmentation protocol by randomly cropping the images to 80-100\% of original sizes and randomly apply horizontal flipping.

\subsubsection{Parameter Setup}

We use ResNet18~\cite{he2016deep} as our backbone. Models are trained with SGD solver, 100 epochs, batch size 128. Learning rate is set to 0.001 and shrinked down to 0.0001 after 80 epochs. Bandwidth $h$ and radius $R$ are fixed at $1$ and $3$, respectively. For photo as source domain, we set $r_0 = 1.5$ and $T = 0.03$. For art or cartoon as source domain, we set $r_0 = 1.5$ and $T = 0.01$. For sketch as source domain, we set $r_0 = 1.2$ and $T = 1.0$.

\subsection{Few-shot Classification}

\subsubsection{Dataset}

We mainly use \emph{mini}-Imagenet~\cite{ravi2016optimization} and CUB~\cite{wah2011caltech} as dataset for few-shot classification. Downloadable links of both dataset can be found in this repository~https://github.com/wyharveychen/CloserLookFewShot.

\emph{mini}-Imagenet contains a subset of 100 classes from the whole ImageNet dataset~\cite{deng2009imagenet} and contains 600 images for each class.Following settings in previous works~\cite{ravi2016optimization}, we randomly divide the whole 100 classes into 64 training classes,16 validation classes and 20 novel classes.

CUB (abbreviation for CUB-200-2011) dataset contains 200 classes with 11788 images.  We divide it into 100 base classes, 50 validation classes and 50 novel classes following~\citet{hilliard2018few}.

We also test our models on the cross-domain scenario, namely $\emph{mini}$-Imagenet$\rightarrow$CUB, where \emph{mini}-ImageNet is used as our base class and the 50 validation and 50 novel classes come from CUB.

Following ~\citet{chen2018a}, we apply data augmentation including random crop, horizontal flip and color jitter.

\subsubsection{Parameter Setup}

We use 4-layer convolutional neural network (Conv-4) as our backbone, following~\cite{snell2017prototypical}. All methods are trained from scratch and use the Adam optimizer with initial learning rate $10^{-3}$. In meta-training stage, we train 60000 episodes for 5-way 5-shot classification without data augmentation, and 80000 episodes for 5-way 1-shot classification without data augmentation. When data augmentation is applied, we add an extra 20000 episodes in meta-training stage. In each episode, we sample 5 classes to form 5-way classification. For each class, we pick k labeled instances as our support set and 16 instances for the query set for a k-shot task. Drop coefficient $r_0$, temperature $T$, bandwidth $h$ and radius $R$ are fixed at $0.1$, $0.5$, $1$ and $3$, respectively. InfoDrop is applied in first two convolutional layers for Conv-4 network, which we use as the backbone through all experiments.

In the fine-tuning or meta-testing stage for all methods, we average the results over 600 experiments. In each experiment, we randomly sample 5 classes from novel classes, and in each class, we also pick k instances for the support set and 16 for the query set. For other settings, we follow the protocol in~\citet{chen2018a}.

Finally, it is worth noting that since we use the re-implementation in~\citet{chen2018a}, results of baseline methods may be higher than reported in their original papers. Please refer to~\citet{chen2018a} for more details.

\subsection{Robustness against Image Corruption}

\subsubsection{Dataset}
\label{robustness_corruption_dataset}

For clean images, we use Caltech-256~\cite{griffin2007caltech} as dataset. It consists of 257 object categories containing a total of 30,607 images with high resolution. Dataset can be downloaded from~\url{http://www.vision.caltech.edu/Image_Datasets/Caltech101/Caltech101.html}. We manually split 20\% of images as the test set. Rescaling and random cropping are used as data augmentation following the protocol in~\citet{he2016deep}.

For generation of corrupted images, we use the library provided in~\citet{hendrycks2019benchmarking}. Original code for corruption generation can be found in \url{https://github.com/hendrycks/robustness/tree/master/ImageNet-C/imagenet_c}. The repository contains 18 types of corruptions: `gaussian noise', `shot noise', `impulse noise', `defocus blur', `motion blur', `zoom blur', `snow', `frost', `fog', `brightness', `contrast', `elastic transform', `pixelate', `jpeg compression', `speckle noise', `gaussian blur', `spatter', `saturate'. The repository provides 5 different levels of corruption severity. In our experiments, we use the highest level, \ie, level-5 severity.

\subsubsection{Parameter Setup}

We train all models for 10 epochs. We use SGD with learning rate 0.01 for 5 epochs, 0.001 for 3 epochs and 0.0001 for 2 epochs. Through all experiments, we only apply InfoDrop to the first convolutional layer before all residual blocks of ResNet18. Bandwidth $h$ and radius $R$ are fixed at $1$ and $3$, respectively. For InfoDrop applied on baseline model, we set $r_0 = 0.7$ and $T = 0.3$. For InfoDrop applied together with adversarial training, we set $r_0 = 1.5$ and $T = 0.03$. For adversarial training, we use 20 runs of PGD attack~\cite{madry2017towards} with $l_\infty$ norm of $1 / 255$. Here we use a relatively small norm to simulate the situation where severity of corruption may exceed the norm of adversarial training. Note that we mainly evaluate InfoDrop's incremental effect on baseline and adversarial methods, while not directly comparing InfoDrop with adversarial training.

\subsection{Adversarial Robustness}

\subsubsection{Dataset}

For evaluation of adversarial robustness, we use two datasets separately, \viz Caltech-256 and CIFAR10. For Caltech-256, as in~\ref{robustness_corruption_dataset}, we manually split 20\% of images as the test set and use rescaling and random cropping for data augmentation. For CIFAR10, we adopt the protocol in~\citet{zhang2019you}.

\subsubsection{Parameter Setup}

For experiments on Caltech-256, we train all models for 10 epochs. We use SGD with learning rate 0.01 for 5 epochs, 0.001 for 3 epochs and 0.0001 for 2 epochs. We apply InfoDrop to the first convolutional layer and first residual block of ResNet18. Bandwidth $h$ and radius $R$ are fixed at $1$ and $3$, respectively. For InfoDrop applied on baseline model, we set $r_0 = 1.5$ and $T = 0.3$. For InfoDrop applied together with adversarial training, we set $r_0 = 2.0$ and $T = 0.01$. For adversarial training, we use 20 runs of PGD attack~\cite{madry2017towards}.

For experiments on CIFAR10, we follow the protocol in~\citet{zhang2019you}. We train models for 105 epochs as a common practice. The learning rate is set to $5e-2$ initially, and is reduced by 10 times at epoch 79, 90 and 100, respectively. We use a batch size of 256, a weight decay of $5e-4$ and a momentum of 0.9 for both algorithm. For adversarial attacks, we use 20 runs of PGD with $l_\infty$ norm of $8 / 255$ and step size of $2 / 255$. We apply InfoDrop only on the first convolutional layer of ResNet18. We set $r_0 = 1.2$, $T = 0.01$, $h = 1$, $R = 3$.



\subsection{Shape-bias of InfoDrop}

In the plotting of CNN's saliency map and experiments of patch shuffling, we use photo-domain in PACS as our dataset and adopt the same settings as in domain generalization. In style transfer, we use pretrained ResNet18 and finetune on content and style images from the repository~https://github.com/xunhuang1995/AdaIN-style.


\end{document}